\pdfoutput=1

\documentclass[11pt]{article}

\usepackage[]{acl}

\usepackage{times}
\usepackage{latexsym}

\usepackage[T1]{fontenc}

\usepackage[utf8]{inputenc}

\usepackage{microtype}

\usepackage{inconsolata}
\usepackage{graphicx}
\usepackage{caption}
\usepackage{subcaption}
\usepackage{diagbox}

\usepackage{tabularx}
\usepackage{multirow}
\usepackage{tablefootnote}
\usepackage{algorithm}
\usepackage{algpseudocode}
\usepackage{enumerate}

\title{Resolving Transcription Ambiguity in Spanish: A Hybrid Acoustic-Lexical System for Punctuation Restoration}

\author{Xiliang Zhu$^*$, \ Chia-Tien Chang$^*$, \ Shayna Gardiner, \ David Rossouw, \ Jonas Robertson \\
        Dialpad Canada Inc.\\
        \texttt{\{xzhu, karol.chang, sgardiner, davidr, jonas\}@dialpad.com} \\}

\begin{document}

\maketitle
\begin{abstract}
Punctuation restoration is a crucial step after Automatic Speech Recognition (ASR) systems to enhance transcript readability and facilitate subsequent NLP tasks. Nevertheless, conventional lexical-based approaches are inadequate for solving the punctuation restoration task in Spanish, where ambiguity can be often found between unpunctuated declaratives and questions. In this study, we propose a novel hybrid acoustic-lexical punctuation restoration system for Spanish transcription, which consolidates acoustic and lexical signals through a modular process. Our experiment results show that the proposed system can effectively improve F1 score of question marks and overall punctuation restoration on both public and internal Spanish conversational datasets. Additionally, benchmark comparison against LLMs (Large Language Model) indicates the superiority of our approach in accuracy, reliability and latency. Furthermore, we demonstrate that the Word Error Rate (WER) of the ASR module also benefits from our proposed system.
\end{abstract}

\def\thefootnote{*}\footnotetext{These authors contributed equally to this work}\def\thefootnote{\arabic{footnote}}

\section{Introduction}
Automatic Speech Recognition (ASR) systems are applied in a variety of industry applications such as voice assistance and conversation analysis. However, typical ASR systems avoid producing punctuation marks in the transcripts, which leads to poor readability and causes ambiguity in the context \cite{readability}. Therefore, a post-processing step to restore punctuation marks in transcripts is critical for speech-based commercial products.

Lexical-based approaches have been extensively studied in punctuation restoration tasks \cite{punc_survey}. One major advantage of using lexical features is the availability of a massive amount of text data that is often well punctuated, such as Wikipedia. Most of the existing work on punctuation restoration focuses on English. Spanish is little studied, although it is the world's second largest mother tongue and even has more native speakers than English. Although a handful of work has addressed Spanish punctuation restoration using BERT-based approaches in recent years \cite{AutoPunct, zhu-etal-2022-punctuation}, one major challenge in restoring punctuation marks for languages like Spanish has not been fully tackled: the rich morphology of Spanish allows speakers to omit subject pronouns and order words in sentences more freely than in English, which forces Spanish speakers to rely more on prosodic features when distinguishing questions from declarative sentences.  These characteristics present a unique challenge from an NLP perspective when written transcripts are the main source of information for models.

In order to address the challenges in predicting Spanish question marks and improve the overall punctuation restoration accuracy, we introduce a hybrid punctuation restoration system leveraging both acoustic and lexical signals for Spanish conversations. While previous work on multimodal methodologies often requires large-scale, parallel audio-text data \cite{7953248}, or additional audio encoding and fusion steps \cite{Multimodal_framework}, our approach employs the conventional modular ASR-NLP setup in industry applications with no additional computational cost. Moreover, our system allows independent training of ASR and NLP modules, eliminating the need for massive parallel training resources. The main contributions of this paper are as follows:
\begin{enumerate}
\item Evaluate the impact of including punctuation in Spanish ASR training data on Word Error Rate (WER).
\item Propose a hybrid system for Spanish punctuation restoration leveraging ASR and NLP sequentially.
\item Demonstrate the effectiveness of our system by achieving up to a 2.1\% relative reduction in Word Error Rate (WER) for the Spanish ASR decoder, improving question mark prediction F1 score by over 4\% absolute, and consequently enhancing overall punctuation restoration accuracy on internal and public datasets from the Linguistic Data Consortium (LDC) \cite{Fisher_Spanish_Transcripts, Fisher_Spanish_Speech}, also outperforming top LLMs (Large Language Model) in terms of accuracy, reliability and latency.

\end{enumerate}

\begin{figure*}[t]
\centering
\includegraphics[width=1\textwidth, height=6cm]{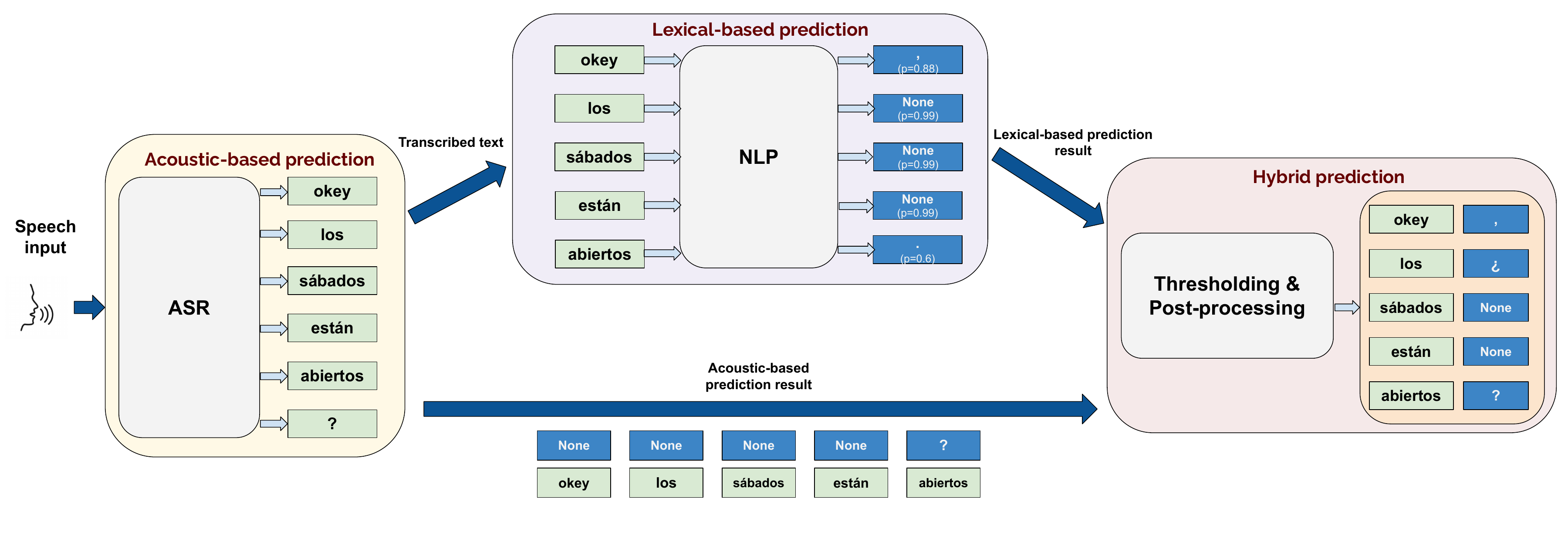}
\caption{\small Overview of our hybrid punctuation restoration system, showing the example of an ambiguous unpunctuated utterance "\textit{okey los sábados están abiertos}" (can be interpreted as "OK, they are open on Saturdays." or "OK, are they open on Saturdays?") processed as "\textit{Okey, ¿los sábados están abiertos?}"}
\label{fig1}
\end{figure*}

\section{Background}

\subsection{Related Work}
\label{sec:2.1}

Punctuation restoration is often formulated as a sequence labeling task, where punctuation marks are predicted at appropriate positions in a sequence of words. Early studies used  lexical-based methods such as n-gram language models \cite{ngram} and Conditional Random Fields (CRF) \cite{crf}. More recently, long short-term memory (LSTM) \cite{LSTM} and pre-trained large language models have been used \cite{LSTM_for_punc, devlin-etal-2019-bert, fu-etal-2021-improving}. Some works \cite{ASR_System_with_Automatic_Punctuation, SPGISpeech} proposed a speech recognition system with direct punctuation output, but it is unclear whether this approach is more effective than a traditional lexicon-based approach.  For Spanish, a multilingual LSTM-based approach was studied in \cite{li20m_interspeech}. \cite{zhu-etal-2022-punctuation} proposed a transformer-based architecture with transfer learning to overcome the Spanish resource limitation and \cite{AutoPunct} integrated silence embedding into BERT; however, as far as we are aware, no study has yet investigated the use of a hybrid approach incorporating acoustic input for the task of Spanish punctuation restoration.

Recent advances in LLMs such as ChatGPT\footnote{\url{https://openai.com/blog/chatgpt}}, GPT4\footnote{\url{https://openai.com/gpt-4}} and PaLM2\footnote{\url{https://blog.google/technology/ai/google-palm-2-ai-large-language-model/}} have reshaped the approaches for many NLP tasks. \cite{qin2023chatgpt} found that ChatGPT performs well on tasks favouring reasoning capabilities while still faces challenges in sequence tagging tasks. \cite{lai2023chatgpt} studied the multilingual capability of ChatGPT and found that it shows less optimal performance compared to task-specific models in different languages. However, the application of LLM in punctuation restoration task has not been studied yet to the best of our knowledge.

\subsection{Ambiguity in Unpunctuated Spanish Text}
\label{sec:2.2}

Identifying Spanish questions from unpunctuated text is a challenging task. There are three relevant sociolinguistic features to consider for question identification in Spanish.

First, declarative sentences can occasionally become questions on intonation and context alone; e.g. {\fontfamily{cmr}\selectfont \textit{ustedes no pueden mandar un cheque con la orden}} can be either a declarative or a question. This is true even for its English counterpart -- both {\fontfamily{cmr}\selectfont\textit{Can’t you send a cheque with the order?}} and {\fontfamily{cmr}\selectfont\textit{You can’t send a cheque with the order?}} are well-formed -- but the phenomenon is extremely common in Spanish \cite{Brown_and_Riveras_2011, RAYMOND201550, Cuza_2016}.  In fact, in Caribbean Spanish, it is becoming increasingly more common to see questions like {\fontfamily{cmr}\selectfont\textit{¿ustedes no pueden mandar un cheque con la orden? (You can’t send a cheque with the order?)}}, which has typical declarative syntax, rather than {\fontfamily{cmr}\selectfont\textit{¿no pueden ustedes mandar un cheque con la orden? (Can’t you send a cheque with the order?)}}, which uses subject-verb inverted order \cite{Brown_and_Riveras_2011}.

Second, Spanish morphosyntax also allows the reverse to occur: that is, declarative sentences can have subject-verb inversion too \cite{Mackenzie_2021}, meaning that the question {\fontfamily{cmr}\selectfont\textit{no pueden ustedes mandar un cheque con la orden}} above is also a perfectly well-formed declarative sentence.

Third, Spanish is a pro-drop language: due to richly-inflected morphology, it is possible to drop a subject pronoun entirely, using a verb’s suffix alone to identify its subject -- and removing the possibility of subject-verb inversion. For instance, the above example could easily become {\fontfamily{cmr}\selectfont\textit{no pueden mandar un cheque con la orden}} or {\fontfamily{cmr}\selectfont\textit{¿no pueden mandar un cheque con la orden?}}. In a small survey of our own data, we reviewed 200 utterances, in which there were 180 questions, of which 125 (69\%) were pro-dropped -- leaving only 55 questions with a fully realized subject noun or pronoun.

These facts make subject-verb inversion a much less helpful tool for definitively identifying questions in Spanish than it is for English, which consequently limits the performance of lexical-based NLP models in the Spanish punctuation restoration task. We also know that Spanish speakers themselves do not rely on lexical information alone to distinguish questions from declaratives: evidence suggests that acoustic features are measurably different when the speaker intends an utterance as a question rather than a declarative, and that this is true across many varieties of Spanish \cite{Face_2005, Willis_2007, Lee_2010, Armstrong_2017}.

\section{Method}
\subsection{System Overview}
\label{sec:3.1}

Our hybrid punctuation restoration system is used in a Spanish call center product. In the real-time system pipeline, the audio from customer support phone calls is first transcribed by the ASR module, then text output is fed into the downstream NLP module. Instead of adding an extra acoustic encoder and combining it with a lexical encoder as proposed in \cite{Multimodal_framework, Unified_Multimodal}, we directly train the ASR decoder to predict target punctuation marks. The ASR punctuation predictions (acoustic-based) are combined with the NLP module predictions (lexical-based) via a probability thresholding process. A heuristic-based post-processing step is then applied to make corrections in the prediction as the final step. Our system is illustrated in Figure \ref{fig1}. The set of Spanish punctuation marks  predicted by the system are: OPEN\_QUESTION (¿)\footnote{An open question mark (¿) is used at the start position of a question in Spanish}, CLOSE\_QUESTION (?), COMMA (,), PERIOD (.), and NONE (for tokens that have no associated punctuation marks).

\subsection{Acoustic-based Prediction}
\label{sec:3.2}
Acoustic features, including intonation and prosody, play an important role in distinguishing declarative and interrogative sentences in Spanish, as described in section \ref{sec:2.2}. In order to leverage our ASR module to directly predict punctuation marks from the speech signal, we keep each target punctuation mark in our ASR training data and treat it as an individual token by separating it from surrounding words; an example of predicting CLOSE\_QUESTION is shown in Figure \ref{fig1}. Note that we omit OPEN\_QUESTION from the training data since it can mostly be restored by heuristics in the following post-processing step.
 
We use an End-to-End based ASR system provided by the Nemo toolkit \cite{kuchaiev2019nemo}. The applied Conformer-CTC architecture is slightly different from the original Conformer architecture \cite{Conformer}, where the LSTM decoder is replaced with a linear decoder. The encoder uses CTC (Connectionist Temporal Classiﬁcation) loss \cite{Graves:06icml} instead of RNNT (RNN-Transducer) \cite{journals/corr/abs-1211-3711} which makes it a non-autoregressive model. For word prediction, we use an in-house streaming decoder with language model shallow fusion.

\subsection{Lexical-based Prediction}
\label{sec:3.3}
The lexical-based approach is capable of predicting all supported punctuation marks outlined in section \ref{sec:3.1}, which consumes unpunctuated transcribed text emitted from the ASR module as shown in Figure \ref{fig1}. For the NLP module utilized in this lexical-based prediction, we follow the similar structure as utilized in \cite{zhu-etal-2022-punctuation} and \cite{fu-etal-2021-improving}, which is a fine-tuned mBERT (multilingual BERT) \cite{devlin-etal-2019-bert} with an additional token classification head. The output of the prediction indicates the appropriate punctuation marks to be attached to the corresponding input token. Additionally, a probability score (illustrated as \(p\) in Figure \ref{fig1}) is also computed for each token using a softmax layer on top of the prediction logits, which reflects the confidence of each predicted punctuation mark in the lexical-based prediction.

\begin{algorithm}
\small
\caption{\small Thresholding algorithm}\label{alg:1}
\begin{algorithmic}
\State \textbf{Input}:
\State \(Pred_a\): acoustic-based prediction 
\State \(Pred_l\): lexical-based prediction 
\State \(P_l\): probability score of \(Pred_l\)  
\State \(T_{question}\) \& \(T_{declarative}\): hyperparameters

\State \textbf{Output}:
\State \(Pred_c\): consolidated prediction
\State \textbf{Where}:
\State \(C\_Q\): CLOSE\_QUESTION
\\
\If{\(Pred_a\) == \(C\_Q\) and \(Pred_l\) in \{\(PERIOD\), \(COMMA\)\}}
\If{$P_l \leq T_{declarative}$}
\State $Pred_c \gets C\_Q$
\Else
\State $Pred_c \gets Pred_l$
\EndIf
\ElsIf{\(Pred_a\) != C\_Q and \(Pred_l\) == C\_Q}
\If{$P_l \leq T_{question}$}
\State $Pred_c \gets PERIOD$
\Else
\State $Pred_c \gets C\_Q$
\EndIf
\Else
\State $Pred_c \gets Pred_l$
\EndIf
\end{algorithmic}
\end{algorithm}

\subsection{Hybrid Prediction}
\label{sec:3.4}
To consolidate the results from both acoustic-based and lexical-based predictions, we introduce a probability thresholding step based on the probability score generated by the lexical-based prediction. Our approach focuses on improving Spanish question prediction, which employs a set of threshold values \(T_{question}\) and \(T_{declarative}\) as hyperparameters. These thresholds represent the minimal probability score the lexical-based prediction needs to have when conflicting with acoustic-based prediction. The detailed thresholding algorithm is illustrated in Algorithm \ref{alg:1}. The optimal values of  \(T_{question}\) and \(T_{declarative}\) are identified through grid search towards the development dataset in our experiment\footnote{We found [0.7, 0.8] is usually a reasonable range to start with for both \(T_{question}\) and \(T_{declarative}\) in our experiments.}.

A heuristic-based post-processing step (details in Appendix \ref{sec:post-proc}) is also applied after probability thresholding to mitigate the error caused by unmatched OPEN\_QUESTION and CLOSE\_QUESTION in the prediction. For example, as illustrated in the hybrid prediction result in Figure \ref{fig1}, an OPEN\_QUESTION is added on the first token of the word chunk {\fontfamily{cmr}\selectfont\textit{los sábados están abiertos?}} after an unmatched CLOSE\_QUESTION is created after the thresholding process.

\section{Experiment}
\subsection{Datasets}
\label{sec:4.1}
We conduct our experiment using a variety of data resources. Since the proposed system is used in our call center product, the in-domain data resource is our internal audio recording and human-annotated transcripts from real customer support calls in Spanish. This internal data resource consists of 50 hours of audio and around 10,000 rows of corresponding transcribed utterances (more statistical detail is available in Appendix \ref{sec:stat}). Apart from our internal dataset, Linguistic Data Consortium (LDC) Spanish Fisher corpora \cite{Fisher_Spanish_Transcripts, Fisher_Spanish_Speech} is also added as a supplementary resource for real-life human conversations, which has approximately 160 hours of audio with 130,000 rows of transcribed utterances from Spanish telephone conversations. Out of both LDC and our internal data, we leave out 10\% and 5\% as test and development sets respectively in our experiments. Note that in order to evaluate the performance of our system on reference transcripts, we leverage Levenshtein distance to align punctuation marks from each ASR hypothesis to reference transcript in acoustic-based prediction during our evaluation process on the test set. 

Additionally, the open-sourced Spanish datasets from Openslr \cite{openslr_Crowdsourcing, mediaspeech2021} and Common-voice \cite{common_voice} are used in ASR training as well, which collectively provide 1200 hours of audio. A subset of 80,000 utterances were also randomly sampled from the Spanish OpenSubtitle corpus \cite{lison-tiedemann-2016-opensubtitles2016} and added as an extra text-only dataset into the NLP module training process to improve the accuracy of lexical-based prediction. All text-based resources are also used in the language model for the in-house streaming ASR decoder.

\subsection{Experiment Setup}
\label{sec:4.2}
For the ASR module, we use the Nemo \cite{kuchaiev2019nemo} Spanish model {\fontfamily{qcr}\selectfont{\normalsize STT\_Es\_Conformer\_CTC\_Large}} as the pre-trained model. The presented model is fine-tuned for 20 epochs, with the Adam optimizer \cite{adam} and no weight decay. The Noam learning scheduler \cite{vaswani2017attention} is used with a warmup of 100 steps and a learning rate of 0.01.

In consideration of the real-time inference speed, we take only the bottom 6 layers of {\fontfamily{qcr}\selectfont{\normalsize bert-base-multilingual-cased}} from Hugging Face \cite{wolf-etal-2020-transformers} library as the backbone of our NLP module. The 6-layer mBERT is then fine-tuned through a token classification task using all lexical training data described in section \ref{sec:4.1}. The NLP module is trained using the Adam optimizer with 4 epochs and a learning rate of 3e-5.

In the subsequent sections, all assessments are performed utilizing a single Intel Xeon 2.20GHz CPU, 1.5G memory and under identical network connection condition.

\begin{table}[t]
\small
\centering
\begin{tabular}{p{1.5cm}p{1cm}p{1.8cm}p{1.8cm}}
\hline
&
Baseline &
ASR w/ C\_Q &
ASR w/ all \\
\hline
Public &
15.77 &
\textbf{15.44 (-2.1\%)} &
16.53 (+4.8\%) \\

Internal &
26.81 &
\textbf{26.36 (-1.7\%)} &
27.95 (+4.3\%) \\
\hline
\end{tabular}

\caption{\small WER and relative changes compared to the baseline on public (LDC) and internal datasets, where the \textit{Baseline} performance is evaluated by the ASR module trained without punctuation. \textit{ASR w/ C\_Q}: ASR module trained with only CLOSE\_QUESTION; \textit{ASR w/ all}: ASR module trained with CLOSE\_QUESTION, PERIOD and COMMA. }
\label{table1}
\end{table}

\begin{table}[t]
\small
\centering
\begin{tabular}{p{3cm}p{1.5cm}p{1.5cm}}
\hline
&
Reliability &
Latency (s) \\
\hline
ChatGPT-few &
92.4\% &
1.13 \\

ChatGPT-zero &
87.9\% &
1.10 \\

PaLM2-few &
28.7\% &
0.56 \\

PaLM2-zero &
28.6\% &
0.49 \\

Our system (excl. ASR) &
- &
\textbf{0.04} \\

\hline
\end{tabular}

\caption{\small Reliability and Latency comparison between LLM APIs (with both zero- and few- shot prompting) and our system (excluding ASR latency), averaged over all internal and public test samples. Latency shown as "seconds per input utterance".}
\label{table2}
\end{table}

\begin{table*}[t]
\small
\centering
\begin{tabularx}{1\textwidth} { 
  | >{\centering\arraybackslash}X
  | >{\centering\arraybackslash}X 
   >{\centering\arraybackslash}X
   >{\centering\arraybackslash}X
   >{\centering\arraybackslash}X
   >{\centering\arraybackslash}X
  | >{\centering\arraybackslash}X 
   >{\centering\arraybackslash}X
   >{\centering\arraybackslash}X
   >{\centering\arraybackslash}X
   >{\centering\arraybackslash}X| }
\hline
&
\multicolumn{5}{c|}{Public (LDC) data} & 
\multicolumn{5}{c|}{Internal data} \\
&
Lexical &
Acoustic &
ChatGPT-zero  &
ChatGPT-few  &
Hybrid  &
Lexical &
Acoustic &
ChatGPT-zero  &
ChatGPT-few  &
Hybrid \\
\hline
C\_Q &
54.0  &
51.5  &
13.5  &
13.9  &
\textbf{58.2}  &
47.3  &
28.2  &
24.4  &
28.6  &
\textbf{51.7}  \\

O\_Q &
50.6  &
- &
11.1 &
11.7  &
\textbf{52.7}  &
44.5  &
- &
22.9  &
25.4  &
\textbf{47.0}  \\

COMMA &
\textbf{60.8}  &
- &
43.5  &
51.0  &
60.7  &
68.9  &
- &
47.2 &
60.4  &
\textbf{69.0}  \\

PERIOD &
87.7  &
- &
58.6 &
58.8 &
\textbf{88.0}  &
83.6  &
-&
59.5 &
72.4 &
\textbf{83.8}  \\

Overall$^1$ &
74.49  &
- &
44.92 &
49.57 &
\textbf{74.83}  &
72.29  &
- &
48.04 &
61.20 &
\textbf{72.61}  \\
\hline
\multicolumn{7}{l}{\small$^1$Micro average of all punctuation}

\end{tabularx}
\caption{\small F1 score comparison over all punctuation marks with different approaches. \textit{C\_Q}: CLOSE\_QUESION; \textit{O\_Q}: OPEN\_QUESTION; \textit{Lexical}: lexical-based prediction; \textit{Acoustic}: acoustic-based prediction; \textit{Hybrid}: our proposed hybrid system with consolidated prediction; \textit{ChatGPT-few/zero}: ChatGPT with few/zero-shot prompting, details in Appendix \ref{sec:appendix}.}
\label{table3}
\end{table*}

\begin{table*}[t]
\small
\centering
\begin{tabularx}{1\textwidth} { 
  | c 
  | >{\centering\arraybackslash}X 
   >{\centering\arraybackslash}X 
   >{\centering\arraybackslash}X
   >{\centering\arraybackslash}X 
  | >{\centering\arraybackslash}X
   >{\centering\arraybackslash}X
   >{\centering\arraybackslash}X
   >{\centering\arraybackslash}X| }
\hline
&
\multicolumn{4}{c|}{Public (LDC) data} &
\multicolumn{4}{c|}{Internal data} \\

&
Lexical &
Acoustic &
Union &
Threshold &
Lexical &
Acoustic &
Union &
Threshold \\
\hline
Precision &
48.0 &
\textbf{65.3}$^2$ &
47.0 &
53.1 &
63.7 &
\textbf{98.4}$^3$ &
66.0 &
66.1 \\
Recall &
61.7 &
44.1 &
\textbf{71.1} &
64.3 &
37.7 &
16.4 &
42.1 &
\textbf{42.4} \\
F1 &
54.0 &
51.5 &
56.5 &
\textbf{58.2} &
47.3 &
28.2 &
51.4 &
\textbf{51.7} \\
\hline
\multicolumn{9}{l}{\small $^2$19.3\% of the True Positive prediction is ambiguous in unpunctuated text, and not identified as questions by Lexical.}\\
\multicolumn{9}{l}{\small$^3$32.3\% of the True Positive prediction is ambiguous in unpunctuated text, and not identified as questions by Lexical.}
\end{tabularx}
\caption{\small F1, precision and recall comparison on \textbf{CLOSE\_QUESTION} using different approaches. \textit{Lexical}: lexical-based prediction; \textit{Acoustic}: acoustic-based prediction; \textit{Union}: the union of CLOSE\_QUESTION predictions from both lexical and acoustic prediction; \textit{Threshold}: our proposed thresholding process to consolidate lexical and acoustic predictions. }
\label{table4}
\end{table*}

\section{Results}
\subsection{Evaluation on Speech Recognition}
\label{sec:5.1}
We first evaluate the performance impact by introducing CLOSE\_QUESTION prediction in our ASR module. Word-Error-Rate (WER) is a standard metric for the ASR system. A lower word error rate shows superior accuracy in speech recognition, compared with a higher word error rate. To accurately determine the word error rate of the ASR module, free from punctuation interference, we exclude all punctuation marks in both the ASR hypothesis and reference transcripts while evaluating.
Table \ref{table1} shows WER on both test sets. Compared to our baseline, the ASR module trained only with CLOSE\_QUESTION shows 2.1\% and 1.7\% WER improvement in public (LDC) and the internal test set respectively, which indicates that our ASR module can learn better acoustic features of Spanish interrogative sentences by keeping CLOSE\_QUESTION in training data. In addition to predicting CLOSE\_QUESTION, we also conduct a second experiment to keep all CLOSE\_QUESTION, COMMA and PERIOD in the ASR module, but this unexpectedly increases the WER by up to 4.8\%, which is not a tolerable performance deterioration for our production use. Therefore, we only focus on CLOSE\_QUESTION prediction from the ASR module in our design.

\subsection{Evaluation on Punctuation Restoration}
\label{sec:5.2}

In order to assess the comprehensive proficiency of our system in restoring Spanish punctuation, we conduct a benchmark test against some leading LLMs available on the market. First, we evaluate and compare the runtime performance of producing Spanish punctuation marks from unpunctuated transcripts between (a) utilizing commercial LLM APIs (ChatGPT and PaLM2) and (b) executing our proposed system. Our evaluation criteria for this analysis include two metrics: (1) \textit{Reliability}: the percentage of the results where the original input words can be extracted without any modification or reordering (except casing changes), to measure the impact of the LLM hallucination or other undesired outcomes. (2) \textit{Latency}: the elapsed time to receive responses from API calls for ChatGPT and PaLM2, as well as the total execution time of our lexical and hybrid prediction (excluding ASR latency), under the same environment setting as stated in section \ref{sec:4.2}. Table \ref{table2} presents the \textit{Reliability} and \textit{Latency} comparison, it is clear that except our proposed system, all LLM APIs exhibit various levels of reliability concerns. Additionally, our system shows much lower latency compared to API calls. It is also noteworthy that \textit{Reliability} of PaLM2 stands at a mere 28\% in both zero- and few- shot prompting, suggesting that it is not suitable for the Spanish punctuation restoration task. Details on the API call setup and prompts are listed in Appendix \ref{sec:api} and \ref{sec:prompt}.

Table \ref{table3} presents the comprehensive F1 score performance of our punctuation restoration system and LLM API\footnote{Only reliable outcomes from ChatGPT are evaluated. PaLM2 is left out in this evaluation as it cannot produce reliable results of a large enough size to establish a meaningful comparison, due to its low \textit{Reliability}.} on both public and internal datasets. Note that we also show the performance of the standalone lexical module which represents the conventional BERT-based lexical-only structure used in recent punctuation restoration studies \cite{zhu-etal-2022-punctuation, fu-etal-2021-improving}.  It is clear that both lexical and hybrid predictions demonstrate a substantial accuracy advantage over ChatGPT. Moreover, the hybrid approach, enhanced by the improvements in CLOSE\_QUESTION of up to 4.4\%, exhibits varied degrees of F1 score improvement for all other punctuation marks after our thresholding and post-processing step outlined in section \ref{sec:3.4}. Consequently, our proposed hybrid system outperforms the lexical-only approach by 0.34\% and 0.32\% absolute in overall F1 score respectively on public and internal datasets. 

To better illustrate the enhancement on CLOSE\_QUESTION from our hybrid system, we additionally provide precision, recall and F1 score details on CLOSE\_QUESTION in Table \ref{table4}. Although with a lower F1 score, acoustic-based prediction exhibits a higher precision in predicting CLOSE\_QUESTION compared to lexical prediction in both testing datasets. In addition, up to 32.3\% of True Positives from the acoustic prediction is ambiguous in unpunctuated text and does not overlap with that in lexical prediction. In order to demonstrate the effectiveness of our proposed thresholding process to consolidate acoustic and lexical predictions as described in section \ref{sec:3.4}, we compare it with a naive union of the two on CLOSE\_QUESTION. Table \ref{table4} shows that \textit{Thresholding} consistently outperforms \textit{Union} in both datasets. As a result, with our thresholding approach, the F1 score for CLOSE\_QUESTION is noticeably improved by 4.2\% and 4.4\% compared to lexical-only prediction across public and internal datasets respectively.

\section{Future Work}
From the evaluation result in section \ref{sec:5.1}, contrary to the WER improvement when predicting only CLOSE\_QUESTION by the ASR module, we discovered a WER deterioration when adding COMMA and PERIOD to the prediction. Future work may focus on establishing a possible cause for this change. In addition, lexical ambiguity between questions and declarations exists beyond Spanish; thus, a natural next step would be evaluating our system in other human languages. 

\section{Conclusion}
In this study, we propose a hybrid acoustic-lexical punctuation restoration system for Spanish conversational transcripts, with a focus to address the ambiguity in unpunctuated Spanish questions. The proposed system leverages an ASR decoder to make direct predictions of Spanish question marks, which are later consolidated with lexical predictions from an NLP module. We evaluate the system on both internal and public datasets and show that it can effectively enhance Spanish question marks prediction, and consequently improve the overall punctuation restoration accuracy. Additional benchmark indicates that our proposed system outperforms some top LLMs in accuracy, latency and reliability. Furthermore, we demonstrate that keeping question marks in the ASR decoder vocabulary results in an improved WER of the ASR module alone.

\section{Ethical Considerations}
During our internal data collection process, we implement a data retention policy for all our users, such that user consent is obtained prior to any data collection. In addition, we have ensured that all the annotators involved in the transcription process of our internal dataset are paid with adequate compensation. Moreover, to protect the privacy and confidentiality of individuals, the dataset underwent further processing to remove any sensitive, personal, or identifiable information. 

\bibliography{anthology,custom}

\begin{thebibliography}{39}
\expandafter\ifx\csname natexlab\endcsname\relax\def\natexlab#1{#1}\fi

\bibitem[{Ardila et~al.(2019)Ardila, Branson, Davis, Henretty, Kohler, Meyer, Morais, Saunders, Tyers, and Weber}]{common_voice}
Rosana Ardila, Megan Branson, Kelly Davis, Michael Henretty, Michael Kohler, Josh Meyer, Reuben Morais, Lindsay Saunders, Francis~M. Tyers, and Gregor Weber. 2019.
\newblock \href {https://doi.org/10.48550/ARXIV.1912.06670} {Common {V}oice: {A} {M}assively-{M}ultilingual {S}peech {C}orpus}.

\bibitem[{Armstrong(2017)}]{Armstrong_2017}
Meghan~E. Armstrong. 2017.
\newblock \href {https://doi.org/doi:10.1515/probus-2014-0016} {Accounting for intonational form and function in puerto rican spanish polar questions}.
\newblock \emph{Probus}, 29(1):1--40.

\bibitem[{Brown and Rivas(2011)}]{Brown_and_Riveras_2011}
Esther Brown and Javier Rivas. 2011.
\newblock \href {https://doi.org/10.1075/sic.8.1.02bro} {Subject-verb word order in spanish interrogatives: A quantitative analysis of puerto rican spanish}.
\newblock \emph{Spanish in Context}, 8.

\bibitem[{Cuza(2016)}]{Cuza_2016}
Alejandro Cuza. 2016.
\newblock \href {https://doi.org/10.1016/j.lingua.2016.04.007} {The status of interrogative subject–verb inversion in spanish-english bilingual children}.
\newblock \emph{Lingua}, 180.

\bibitem[{Devlin et~al.(2019)Devlin, Chang, Lee, and Toutanova}]{devlin-etal-2019-bert}
Jacob Devlin, Ming-Wei Chang, Kenton Lee, and Kristina Toutanova. 2019.
\newblock \href {https://doi.org/10.18653/v1/N19-1423} {{BERT}: Pre-training of deep bidirectional transformers for language understanding}.
\newblock In \emph{Proceedings of the 2019 Conference of the North {A}merican Chapter of the Association for Computational Linguistics: Human Language Technologies, Volume 1 (Long and Short Papers)}, pages 4171--4186, Minneapolis, Minnesota. Association for Computational Linguistics.

\bibitem[{Face(2005)}]{Face_2005}
Timothy~L. Face. 2005.
\newblock \href {https://www.jstor.org/stable/41678113} {F0 peak height and the perception of sentence type in castilian spanish}.
\newblock \emph{Revista Internacional de Lingüística Iberoamericana}, 3:49--65.

\bibitem[{Fu et~al.(2021)Fu, Chen, Laskar, Bhushan, and Corston-Oliver}]{fu-etal-2021-improving}
Xue-Yong Fu, Cheng Chen, Md~Tahmid~Rahman Laskar, Shashi Bhushan, and Simon Corston-Oliver. 2021.
\newblock \href {https://doi.org/10.18653/v1/2021.wnut-1.19} {Improving punctuation restoration for speech transcripts via external data}.
\newblock In \emph{Proceedings of the Seventh Workshop on Noisy User-generated Text (W-NUT 2021)}, pages 168--174, Online. Association for Computational Linguistics.

\bibitem[{González-Docasal et~al.(2021)González-Docasal, García-Pablos, Arzelus, and Álvarez}]{AutoPunct}
Ander González-Docasal, Aitor García-Pablos, Haritz Arzelus, and Aitor Álvarez. 2021.
\newblock \href {http://journal.sepln.org/sepln/ojs/ojs/index.php/pln/article/view/6377} {Autopunct: A {BERT}-based {A}utomatic {P}unctuation and {C}apitalisation {S}ystem for {S}panish and {B}asque}.
\newblock \emph{Procesamiento del Lenguaje Natural}, 67(0):59--68.

\bibitem[{Graff et~al.(2010{\natexlab{a}})Graff, Huang, Cartagena, Walker, and Cieri}]{Fisher_Spanish_Transcripts}
David Graff, Shudong Huang, Ingrid Cartagena, Kevin Walker, and Christopher Cieri. 2010{\natexlab{a}}.
\newblock \href {https://doi.org/10.35111/s30q-sn19} {Fisher {S}panish - {T}ranscripts {LDC}2010{T}04}.
\newblock Web Download. Philadelphia: Linguistic Data Consortium.

\bibitem[{Graff et~al.(2010{\natexlab{b}})Graff, Huang, Cartagena, Walker, and Cieri}]{Fisher_Spanish_Speech}
David Graff, Shudong Huang, Ingrid Cartagena, Kevin Walker, and Christopher Cieri. 2010{\natexlab{b}}.
\newblock \href {https://doi.org/10.35111/skrw-t863} {{F}isher {S}panish {S}peech {LDC}2010{S}01}.
\newblock Web Download. Philadelphia: Linguistic Data Consortium.

\bibitem[{Gravano et~al.(2009)Gravano, Jansche, and Bacchiani}]{ngram}
Agustin Gravano, Martin Jansche, and Michiel Bacchiani. 2009.
\newblock \href {https://doi.org/10.1109/ICASSP.2009.4960690} {Restoring punctuation and capitalization in transcribed speech}.
\newblock In \emph{2009 IEEE International Conference on Acoustics, Speech and Signal Processing}, pages 4741--4744.

\bibitem[{Graves et~al.(2006)Graves, Fernandez, Gomez, and Schmidhuber}]{Graves:06icml}
A.~Graves, S.~Fernandez, F.~Gomez, and J.~Schmidhuber. 2006.
\newblock Connectionist temporal classification: Labelling unsegmented sequence data with recurrent neural nets.
\newblock In \emph{ICML '06: Proceedings of the International Conference on Machine Learning}.

\bibitem[{Graves(2012)}]{journals/corr/abs-1211-3711}
Alex Graves. 2012.
\newblock \href {http://dblp.uni-trier.de/db/journals/corr/corr1211.html#abs-1211-3711} {Sequence transduction with recurrent neural networks}.
\newblock \emph{CoRR}, abs/1211.3711.

\bibitem[{Guan(2020)}]{ASR_System_with_Automatic_Punctuation}
Yushi Guan. 2020.
\newblock \href {https://doi.org/10.48550/ARXIV.2012.02012} {End to {E}nd {ASR} {S}ystem with {A}utomatic {P}unctuation {I}nsertion}.

\bibitem[{Guevara-Rukoz et~al.(2020)Guevara-Rukoz, Demirsahin, He, Chu, Sarin, Pipatsrisawat, Gutkin, Butryna, and Kjartansson}]{openslr_Crowdsourcing}
Adriana Guevara-Rukoz, Isin Demirsahin, Fei He, Shan-Hui~Cathy Chu, Supheakmungkol Sarin, Knot Pipatsrisawat, Alexander Gutkin, Alena Butryna, and Oddur Kjartansson. 2020.
\newblock \href {https://www.aclweb.org/anthology/2020.lrec-1.801} {{Crowdsourcing {L}atin {A}merican {S}panish for {L}ow-{R}esource {T}ext-to-{S}peech}}.
\newblock In \emph{Proceedings of The 12th Language Resources and Evaluation Conference (LREC)}, pages 6504--6513, Marseille, France. European Language Resources Association (ELRA).

\bibitem[{Gulati et~al.(2020)Gulati, Qin, Chiu, Parmar, Zhang, Yu, Han, Wang, Zhang, Wu, and Pang}]{Conformer}
Anmol Gulati, James Qin, Chung-Cheng Chiu, Niki Parmar, Yu~Zhang, Jiahui Yu, Wei Han, Shibo Wang, Zhengdong Zhang, Yonghui Wu, and Ruoming Pang. 2020.
\newblock \href {https://doi.org/10.48550/ARXIV.2005.08100} {Conformer: {C}onvolution-augmented {T}ransformer for {S}peech {R}ecognition}.

\bibitem[{Hochreiter and Schmidhuber(1997)}]{LSTM}
Sepp Hochreiter and Jürgen Schmidhuber. 1997.
\newblock \href {https://doi.org/10.1162/neco.1997.9.8.1735} {Long {S}hort-term {M}emory}.
\newblock \emph{Neural computation}, 9:1735--80.

\bibitem[{Jones et~al.(2003)Jones, Wolf, Gibson, Williams, Fedorenko, Reynolds, and Zissman}]{readability}
Douglas Jones, Florian Wolf, Edward Gibson, Elliott Williams, Evelina Fedorenko, Douglas Reynolds, and Marc Zissman. 2003.
\newblock \href {https://doi.org/10.21437/Eurospeech.2003-463} {Measuring the readability of automatic speech-to-text transcripts}.

\bibitem[{Kingma and Ba(2014)}]{adam}
Diederik~P. Kingma and Jimmy Ba. 2014.
\newblock \href {https://doi.org/10.48550/ARXIV.1412.6980} {Adam: {A} {M}ethod for {S}tochastic {O}ptimization}.

\bibitem[{Klejch et~al.(2017)Klejch, Bell, and Renals}]{7953248}
Ondřej Klejch, Peter Bell, and Steve Renals. 2017.
\newblock \href {https://doi.org/10.1109/ICASSP.2017.7953248} {Sequence-to-sequence models for punctuated transcription combining lexical and acoustic features}.
\newblock In \emph{2017 IEEE International Conference on Acoustics, Speech and Signal Processing (ICASSP)}, pages 5700--5704.

\bibitem[{Kolobov et~al.(2021)Kolobov, Okhapkina, Olga~Omelchishina, Bedyakin, Moshkin, Menshikov, and Mikhaylovskiy}]{mediaspeech2021}
Rostislav Kolobov, Olga Okhapkina, Andrey~Platunov Olga~Omelchishina, Roman Bedyakin, Vyacheslav Moshkin, Dmitry Menshikov, and Nikolay Mikhaylovskiy. 2021.
\newblock \href {http://arxiv.org/abs/2103.16193} {{M}edia{S}peech: {M}ultilanguage {ASR} {B}enchmark and {D}ataset}.

\bibitem[{Kuchaiev et~al.(2019)Kuchaiev, Li, Nguyen, Hrinchuk, Leary, Ginsburg, Kriman, Beliaev, Lavrukhin, Cook et~al.}]{kuchaiev2019nemo}
Oleksii Kuchaiev, Jason Li, Huyen Nguyen, Oleksii Hrinchuk, Ryan Leary, Boris Ginsburg, Samuel Kriman, Stanislav Beliaev, Vitaly Lavrukhin, Jack Cook, et~al. 2019.
\newblock Nemo: a toolkit for building ai applications using neural modules.
\newblock \emph{arXiv preprint arXiv:1909.09577}.

\bibitem[{Lai et~al.(2023)Lai, Ngo, Veyseh, Man, Dernoncourt, Bui, and Nguyen}]{lai2023chatgpt}
Viet~Dac Lai, Nghia~Trung Ngo, Amir Pouran~Ben Veyseh, Hieu Man, Franck Dernoncourt, Trung Bui, and Thien~Huu Nguyen. 2023.
\newblock \href {http://arxiv.org/abs/2304.05613} {Chatgpt beyond english: Towards a comprehensive evaluation of large language models in multilingual learning}.

\bibitem[{Lee and M.A.(2010)}]{Lee_2010}
Su~Ar Lee and B.A.and M.A. 2010.
\newblock Absolute interrogative intonation patterns in {B}uenos {A}ires {S}panish.

\bibitem[{Li and Lin(2020)}]{li20m_interspeech}
Xinxing Li and Edward Lin. 2020.
\newblock \href {https://doi.org/10.21437/Interspeech.2020-2052} {{A 43 Language Multilingual Punctuation Prediction Neural Network Model}}.
\newblock In \emph{Proc. Interspeech 2020}, pages 1067--1071.

\bibitem[{Lison and Tiedemann(2016)}]{lison-tiedemann-2016-opensubtitles2016}
Pierre Lison and J{\"o}rg Tiedemann. 2016.
\newblock \href {https://aclanthology.org/L16-1147} {{O}pen{S}ubtitles2016: Extracting large parallel corpora from movie and {TV} subtitles}.
\newblock In \emph{Proceedings of the Tenth International Conference on Language Resources and Evaluation ({LREC}'16)}, pages 923--929, Portoro{\v{z}}, Slovenia. European Language Resources Association (ELRA).

\bibitem[{Lu and Ng(2010)}]{crf}
Wei Lu and Hwee~Tou Ng. 2010.
\newblock \href {https://aclanthology.org/D10-1018} {{B}etter {P}unctuation {P}rediction with {D}ynamic {C}onditional {R}andom {F}ields}.
\newblock In \emph{Proceedings of the 2010 Conference on Empirical Methods in Natural Language Processing}, pages 177--186, Cambridge, MA. Association for Computational Linguistics.

\bibitem[{Mackenzie(2021)}]{Mackenzie_2021}
Ian Mackenzie. 2021.
\newblock \href {https://www.staff.ncl.ac.uk/i.e.mackenzie/wordord.htm} {The linguistics of spanish}.

\bibitem[{O'Neill et~al.(2021)O'Neill, Lavrukhin, Majumdar, Noroozi, Zhang, Kuchaiev, Balam, Dovzhenko, Freyberg, Shulman, Ginsburg, Watanabe, and Kucsko}]{SPGISpeech}
Patrick~K. O'Neill, Vitaly Lavrukhin, Somshubra Majumdar, Vahid Noroozi, Yuekai Zhang, Oleksii Kuchaiev, Jagadeesh Balam, Yuliya Dovzhenko, Keenan Freyberg, Michael~D. Shulman, Boris Ginsburg, Shinji Watanabe, and Georg Kucsko. 2021.
\newblock \href {https://doi.org/10.48550/ARXIV.2104.02014} {{SPGIS}peech: 5,000 hours of transcribed financial audio for fully formatted end-to-end speech recognition}.

\bibitem[{Păiș and Tufiș(2021)}]{punc_survey}
Vasile Păiș and Dan Tufiș. 2021.
\newblock Capitalization and punctuation restoration: a survey.
\newblock \emph{Artificial Intelligence Review}, 55:1681 -- 1722.

\bibitem[{Qin et~al.(2023)Qin, Zhang, Zhang, Chen, Yasunaga, and Yang}]{qin2023chatgpt}
Chengwei Qin, Aston Zhang, Zhuosheng Zhang, Jiaao Chen, Michihiro Yasunaga, and Diyi Yang. 2023.
\newblock \href {http://arxiv.org/abs/2302.06476} {Is chatgpt a general-purpose natural language processing task solver?}

\bibitem[{Raymond(2015)}]{RAYMOND201550}
Chase~Wesley Raymond. 2015.
\newblock \href {https://doi.org/https://doi.org/10.1016/j.langcom.2015.02.001} {Questions and responses in spanish monolingual and spanish–english bilingual conversation}.
\newblock \emph{Language and Communication}, 42:50--68.

\bibitem[{Sunkara et~al.(2020)Sunkara, Ronanki, Bekal, Bodapati, and Kirchhoff}]{Multimodal_framework}
Monica Sunkara, Srikanth Ronanki, Dhanush Bekal, Sravan Bodapati, and Katrin Kirchhoff. 2020.
\newblock \href {https://doi.org/10.48550/ARXIV.2008.00702} {Multimodal {S}emi-supervised {L}earning {F}ramework for {P}unctuation {P}rediction in {C}onversational {S}peech}.

\bibitem[{Vaswani et~al.(2017)Vaswani, Shazeer, Parmar, Uszkoreit, Jones, Gomez, Kaiser, and Polosukhin}]{vaswani2017attention}
Ashish Vaswani, Noam Shazeer, Niki Parmar, Jakob Uszkoreit, Llion Jones, Aidan~N Gomez, {\L}ukasz Kaiser, and Illia Polosukhin. 2017.
\newblock Attention is all you need.
\newblock In \emph{Advances in Neural Information Processing Systems}, pages 5998--6008.

\bibitem[{Willis(2007)}]{Willis_2007}
Erik Willis. 2007.
\newblock \href {https://doi.org/10.5334/jpl.149} {Utterance signaling and tonal levels in dominican spanish declaratives and interrogatives}.
\newblock \emph{Journal of Portuguese Linguistics}, 6:179.

\bibitem[{Wolf et~al.(2020)Wolf, Debut, Sanh, Chaumond, Delangue, Moi, Cistac, Rault, Louf, Funtowicz, Davison, Shleifer, von Platen, Ma, Jernite, Plu, Xu, Le~Scao, Gugger, Drame, Lhoest, and Rush}]{wolf-etal-2020-transformers}
Thomas Wolf, Lysandre Debut, Victor Sanh, Julien Chaumond, Clement Delangue, Anthony Moi, Pierric Cistac, Tim Rault, Remi Louf, Morgan Funtowicz, Joe Davison, Sam Shleifer, Patrick von Platen, Clara Ma, Yacine Jernite, Julien Plu, Canwen Xu, Teven Le~Scao, Sylvain Gugger, Mariama Drame, Quentin Lhoest, and Alexander Rush. 2020.
\newblock \href {https://doi.org/10.18653/v1/2020.emnlp-demos.6} {Transformers: State-of-the-art natural language processing}.
\newblock In \emph{Proceedings of the 2020 Conference on Empirical Methods in Natural Language Processing: System Demonstrations}, pages 38--45, Online. Association for Computational Linguistics.

\bibitem[{Xu et~al.(2016)Xu, Xie, and Yao}]{LSTM_for_punc}
Kaituo Xu, Lei Xie, and Kaisheng Yao. 2016.
\newblock \href {https://doi.org/10.1109/ISCSLP.2016.7918492} {Investigating {LSTM} for punctuation prediction}.
\newblock In \emph{2016 10th International Symposium on Chinese Spoken Language Processing (ISCSLP)}, pages 1--5.

\bibitem[{Zhu et~al.(2022{\natexlab{a}})Zhu, Gardiner, Rossouw, Rold{\'a}n, and Corston-Oliver}]{zhu-etal-2022-punctuation}
Xiliang Zhu, Shayna Gardiner, David Rossouw, Tere Rold{\'a}n, and Simon Corston-Oliver. 2022{\natexlab{a}}.
\newblock \href {https://doi.org/10.18653/v1/2022.deeplo-1.9} {Punctuation restoration in {S}panish customer support transcripts using transfer learning}.
\newblock In \emph{Proceedings of the Third Workshop on Deep Learning for Low-Resource Natural Language Processing}, pages 80--89, Hybrid. Association for Computational Linguistics.

\bibitem[{Zhu et~al.(2022{\natexlab{b}})Zhu, Wu, Cheng, and Wang}]{Unified_Multimodal}
Yaoming Zhu, Liwei Wu, Shanbo Cheng, and Mingxuan Wang. 2022{\natexlab{b}}.
\newblock \href {https://doi.org/10.48550/ARXIV.2202.00468} {Unified {M}ultimodal {P}unctuation {R}estoration {F}ramework for {Mi}xed-{M}odality {C}orpus}.

\end{thebibliography}

\appendix

\section{Appendix}
\label{sec:appendix}

\subsection{Heuristic-based post-processing}
\label{sec:post-proc}
As mentioned in 3.4, we apply the following heuristic-based steps to post-process the prediction result:
\begin{enumerate}
\item Convert all unmatched OPEN\_QUESTION to NONE in the prediction.
\item For all unmatched CLOSE\_QUESTION, change the prediction of the first token in the continuous word chunk (the longest continuous word sequence where no punctuation is predicted in-between) to OPEN\_QUESTION. 
\end{enumerate}

\subsection{Description of our internal dataset}
\label{sec:stat}
Our internal data is collected from audio recordings of Spanish customer support conversations, covering a large range of domains such as retail, technology, automotive and professional services. Our primary focus lies within the North American region, including both Mexican and American accents. The audio duration of the dataset totals around 50 hours. For ASR training purposes, each individual audio clip is broken down into segments based on audio silence, with a maximum of 2 minutes and averaging approximately 18 seconds. Audio clips are also transcribed by the annotators to create text data for NLP training. We provide the statistical summary on the length of the transcribed utterances in Table \ref{table5}.

\begin{table}[ht]
\small
\centering
\begin{tabular}{cccccc}
\hline
&
mean &
medium &
min &
max &
std \\
\hline
num of words &
43.4 &
38.0 &
1.0 &
231.0 &
25.4 \\

\hline
\end{tabular}

\caption{\small Statistical summary on length of the utterances in our internal dataset.}
\label{table5}
\end{table}

\subsection{API call setup}
\label{sec:api}
We use {\fontfamily{qcr}\selectfont gpt-3.5-turbo} for ChatGPT and {\fontfamily{qcr}\selectfont text-bison@001} for PaLM2 in API calls. For both models, \textit{temperature} is set as 0.2 while the maximum token length of output (named as \textit{max\_tokens} in ChatGPT and \textit{maxOutputTokens} in PaLM2) is configured as 1024.

\subsection{Prompt}
\label{sec:prompt}

The following prompts are used in our experiments when calling LLM APIs:
\\

Few-shot prompting:\\
{\fontfamily{qcr}\selectfont{\normalsize \small Without any explanation or modification, add punctuation to the following Spanish transcript from human conversations, use only punctuation marks from this list: comma(,), period(.), open\_question(¿) and close\_question(?).
    Return the punctuated utterance only. 
    Here are some examples:\\
    \#\#\# Input: \{Unpunctuated Spanish Utterance 1\}\\
    \#\#\# Output: \{Punctuated Spanish Utterance 1\} \\
    \\
    \#\#\# Input: \{Unpunctuated Spanish Utterance 2\}\\
    \#\#\# Output: \{Punctuated Spanish Utterance 2\}\\
    \\
    \#\#\# Input: \{Unpunctuated Spanish Utterance 3\}\\
    \#\#\#   Output: \{Punctuated Spanish Utterance 3\}\\
    \\
    Now, add punctuation marks to: \\
    \#\#\# Input: \{\textbf{text}\} \\
    \#\#\# Output:
    }}\\

Zero-shot prompting:\\
{\fontfamily{qcr}\selectfont{\normalsize \small Without any explanation or modification, add punctuation to the following Spanish transcript from human conversations, use only punctuation marks from this list: comma(,), period(.), open\_question(¿) and close\_question(?). Return the punctuated utterance only. 

    Add punctuation marks to:\\
    \#\#\# Input: \{\textbf{text}\} \\
    \#\#\# Output:
    }}\\

where we put the unpunctuated test utterance in the \textbf{text} field. Note that in all of our experiments, we use three in-context examples for few-shot prompting. In addition, we make sure to sample utterances with presence of all targeted punctuation marks in these three in-context examples. Note that both zero-shot and few-shot prompting are used in the evaluation results as presented in Table \ref{table3} and Table \ref{table4}.

\end{document}